\documentclass{article}

\usepackage{graphicx}

\usepackage{bbm}
\usepackage[utf8]{inputenc}
\usepackage[english]{babel}
\usepackage{amsthm}
\usepackage{amsmath}
\usepackage{amsfonts}
\usepackage{booktabs}
\usepackage{xcolor}
\usepackage{subcaption}

 \usepackage{hyperref}
 
 \usepackage[accepted]{icml2019}

\newtheorem*{theorem*}{Theorem}
\newtheorem{lemma}{Lemma}
\newtheorem{definition}{Definition}

\newcommand{\M}{\mathcal{M}}

\newcommand{\E}{\mathbb{E}}

\newcommand{\R}{\mathbb{R}}
\newcommand{\D}{\mathcal{D}}
\newcommand{\A}{\mathcal{A}}
\newcommand{\T}{\mathcal{T}}

\newcommand{\C}{\mathcal{C}}
\newcommand{\K}{\mathcal{K}}
\newcommand{\s}{\mathcal{S}}

\newcommand{\pth}[1]{\left( #1 \right) }
\newcommand{\bpth}[1]{\left[ #1 \right] }
\newcommand{\abs}[1]{{\left| #1 \right| }}
\newcommand{\braces}[1]{\left\{ #1 \right\} }
\newcommand{\cmnt}[1]{\ignorespaces}
\newcommand{\norm}[1]{\left\lVert#1\right\rVert}

\newcommand{\indicator}[1]{\mathbbm{1}_{\braces{#1}}}
\newcommand\given[1][]{\:#1\vert\:}

\icmltitlerunning{The Natural Language of Actions}
\begin{document}
\twocolumn[
\icmltitle{The Natural Language of Actions}
\icmlsetsymbol{equal}{*}
        
\begin{icmlauthorlist}
\icmlauthor{Guy Tennenholtz}{to}
\icmlauthor{Shie Mannor}{to}
\end{icmlauthorlist}

\icmlaffiliation{to}{Faculty of Electrical Engineering, Technion Institute of Technology, Israel}

\icmlcorrespondingauthor{Guy Tennenholtz}{guytenn@gmail.com}
\icmlcorrespondingauthor{Shie Mannor}{shie@technion.ac.il}

\icmlkeywords{Reinforcement Learning, Action Embeddings, Action Representation, Act2Vec}

\vskip 0.3in
]



\printAffiliationsAndNotice{} 

\interfootnotelinepenalty=10000
\begin{abstract} 
We introduce Act2Vec, a general framework for learning context-based action representation for Reinforcement Learning. Representing actions in a vector space help reinforcement learning algorithms achieve better performance by grouping similar actions and utilizing relations between different actions. We show how prior knowledge of an environment can be extracted from demonstrations and injected into action vector representations that encode natural compatible behavior. We then use these for augmenting state representations as well as improving function approximation of Q-values. We visualize and test action embeddings in three domains including a drawing task, a high dimensional navigation task, and the large action space domain of StarCraft II.

\end{abstract}

\section{Introduction}

\begin{figure}[t!]
\vskip 0.2in
\begin{center}
\includegraphics[width=0.25\textwidth]{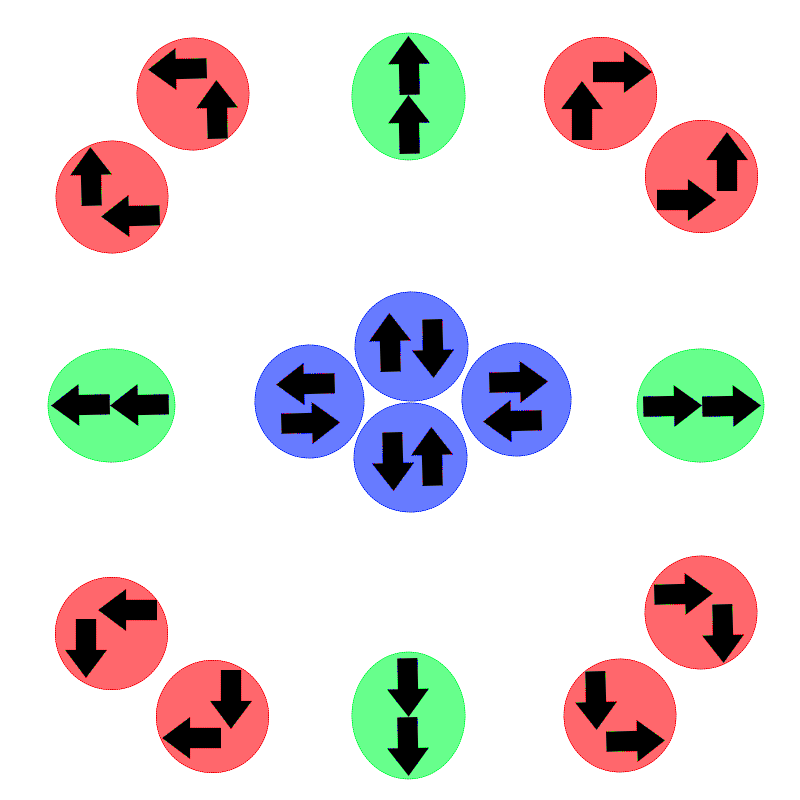}
\caption{A schematic visualization of action pair embeddings in a navigation domain using a distributional representation method. Each circle represents an action of two consecutive movements in the world. Actions that are close have similar contexts. Relations between actions in the vector space have interpretations in the physical world.}
\label{intuition}
\end{center}
\vskip -0.2in
\end{figure}

The question ``What is language" has had implications in the fields of neuropsychology, linguistics, and philosophy. One definition tells us that language is ``a purely human and non-instinctive method of communicating ideas, emotions, and desires by means of voluntarily produced symbols" \cite{language_intro}. Much like humans adopt languages to communicate, their interaction with an environment uses sophisticated languages to convey information. Inspired by this conceptual analogy, we adopt existing methods in natural language processing (NLP) to gain a deeper understanding of the ``natural language of actions" with the ultimate goal of solving reinforcement learning (RL) tasks.
 
In recent years, many advances were made in the field of distributed representations of words \cite{word2vec, glove, ngram2vec}. \cmnt{The main idea of this approach is encoding and storing information about an item within a system by establishing its interactions with other members \cite{word_contex_idea}.} Distributional methods make use of the hypothesis that words which occur in a similar context tend to have similar meaning \cite{word_contex_idea}, i.e., the meaning of a word can be inferred from the distribution of words around it. For this reason, these methods are called ``distributional" methods. Similarly, the context in which an action is executed holds vital information (e.g., prior knowledge) of the environment. This information can be transferred to a learning agent through distributed action representations in a manifold of diverse environments.

\cmnt{
NLP encompasses a wide variety of topics which involve the computational processing and understanding of natural human languages. In recent years, many advances were made in the field of distributed representations of words \cite{word2vec, glove, ngram2vec} . The main idea of this approach is encoding and storing information about an item within a system by establishing its interactions with other members \cite{word_contex_idea}. Word embeddings claim to capture meanings of words, following the Distributional Hypothesis of Meaning \cite{embeddings_meaning}.  Recently, advances in language models managed to outperform traditional models on various tasks \cite{attention_LM, conv_LM}. These advances are readily available for use to construct action representations in a manifold of diverse natural environments.}

In this paper, actions are represented and characterized by the company they keep (i.e., their context). We assume the context in which actions reside is induced by a demonstrator policy (or set of policies). We use the celebrated continuous skip-gram model \cite{skip_gram} to learn high-quality vector representations of actions from large amounts of demonstrated trajectories. In this approach, each action or sequence of actions is embedded in a $d$-dimensional vector that characterizes knowledge of acceptable behavior in the environment.

\begin{figure}[t!]
\vskip 0.2in
\begin{center}
\includegraphics[width=0.47\textwidth]{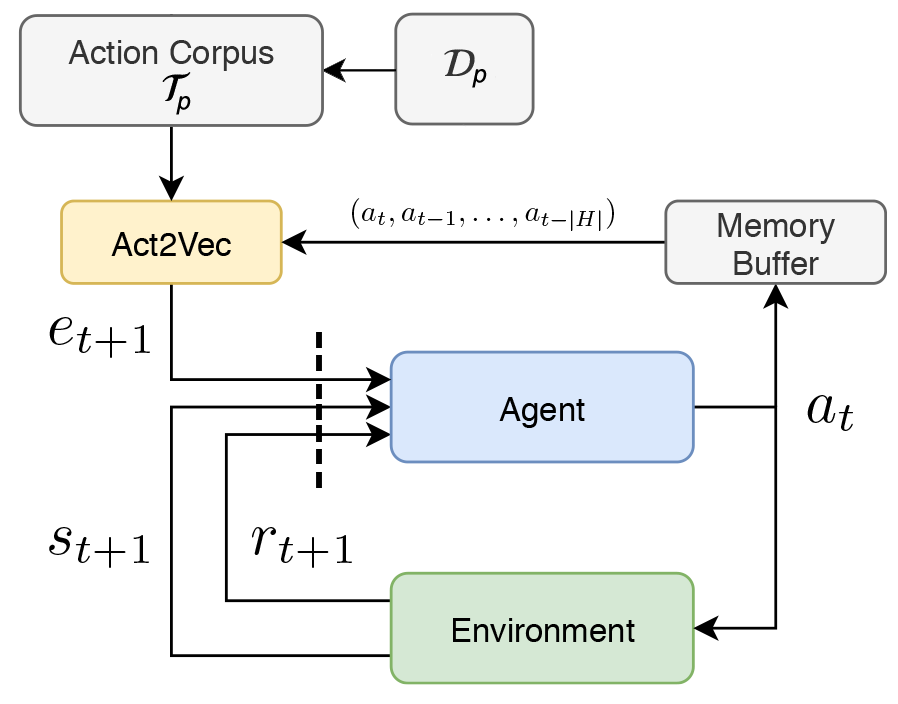}
\vskip -0.1in
\caption{A schematic use of action embedding for state augmentation. Act2Vec is trained over sequences of length $|H|$ actions taken from trajectories in the action corpus $\T_{p}$, sampled from $\D_{p}$. Representation of action histories ($e_{t+1}$) are joined with the state ($s_{t+1}$). The agent $\pi$ thus maps state and action histories $(s_t, a_{t-1}, \hdots a_{t-|H|})$ to an action $a_t$. }
\label{fig:rl_diagram}
\end{center}
\vskip -0.3in
\end{figure}

As motivation, consider the problem of navigating a robot. In its basic form, the robot must select a series of primitive actions of going straight, backwards, left, or right. By reading the words ``straight", ``backwards", ``left", and ``right" the reader already has a clear understanding of their implication in the physical world. Particularly, it is presumed that moving straight and then left should have higher correlation to moving left and then straight, but for the most part contrast to moving right and then backwards. Moreover, it is well understood that a navigation solution should rarely take the actions straight and backwards successively, as it would postpone the arrival to the desired location. \mbox{Figure \ref{intuition}} shows a 2-dimensional schematic of these action pairs in an illustrated vector space. We later show (Section \ref{sec:navigation}) that similarities, relations, and symmetries relating to such actions are present in context-based action embeddings.

Action embeddings are used to mitigate an agent's learning process on several fronts. First, action representations can be used to improve expressiveness of state representations, sometimes even replacing them altogether (Section \ref{sec:drawing}). In this case, the policy space is augmented to include action-histories. These policies map current state and action histories to actions, i.e., $\pi : \s \times \A^\abs{H} \to \A$. Through vector action representations, these policies can be efficiently learned, improving an agent's overall performance. A conceptual diagram presenting this approach is depicted in Figure \ref{fig:rl_diagram}. Second, similarity between actions can be leveraged to decrease redundant exploration through grouping of actions. In this paper, we show how similarity between actions can improve approximation of $Q$-values, as well as devise a cluster-based exploration strategy for efficient exploration in large action space domains \mbox{(Sections \ref{sec:func_approx}, \ref{sec:navigation})}.  

\cmnt{
To this end, action representations seem primarily useful in large action space domains. Nevertheless,  any domain can be expressed through an enlarged action space consisting of all possible sequences of actions of length $k$, $(a_1, \hdots, a_k)$. These sequences of actions produce an augmented space of actions of cardinality $\abs{A}^k$. While there is no apparent reason to produce such a large action space, natural action embeddings can drastically decrease its complexity through similarity of action sequences, as well as sequences that are never permissible. Doing so can decrease the search horizon for solving a given task. Such a process is especially useful in deterministic settings, where the generality of the problem is not affected \footnote{In a deterministic setting, the optimal policy that selects an action $a \in A$ at every time step is equivalent to the optimal policy that selects a sequence of $k$ actions $(a_1, \hdots, a_k) \in A^k$ every $k$ time steps.}.}

Our main contributions in this paper are as follows. (1) We generalize the use of context-based embedding for actions. We show that novel insights can be acquired, portraying our knowledge of the world via similarity, symmetry, and relations in the action embedding space. (2) We offer uses of action representations for state representation, function approximation, and exploration. We demonstrate the advantage in performance on drawing and navigation domains.

This paper is organized as follows. Section \ref{sec:setting} describes the general setting of action embeddings. We describe the Skip Gram with Negative Sampling (SGNS) model used for representing actions and its relation to the pointwise mutual information (PMI) of actions with their context. In \mbox{Section \ref{sec:func_approx}} we show how embeddings can be used for approximating $Q$-values. We introduce a cluster based exploration procedure for large action spaces. Section \ref{sec:empirical} includes empirical uses of action representations for solving reinforcement learning tasks, including drawing a square and a navigation task. We visualize semantic relations of actions in the said domains as well as the large action space domain of \mbox{Starcraft II}. We conclude the paper with related work (Section \ref{sec:related_work}) and a short discussion on future directions \mbox{(Section \ref{sec:discussion}).}

\section{General Setting}
\label{sec:setting}

In this section we describe the general framework of Act2Vec. We begin by defining the context assumptions from which we learn action vectors. We then illustrate that embedding actions based on their point-wise mutual information (PMI) with their contexts contributes favorable characteristics. One beneficial property of PMI-based embedding is that close actions issue similar outcomes. In Section \ref{sec:empirical} we visualize Act2Vec embeddings of several domains, showing their inherent structure is aligned with our prior understanding of the tested domains.

A Markov Decision Processes (MDP) is defined by a 5-tuple $\M_R = (\s, \A, P, R, \gamma)$, where $\s$ is a set of states, $\A$ is a discrete set of actions, ${P:\s\times\A\times\s \to [0,1]}$ is a set of transition probabilities, ${R:\s \to [0,1]}$ is a scalar reward, and $\gamma \in (0,1)$ is a discount factor. We consider the general framework of multi-task RL in which $R$ is sampled from a task distribution $\D_R$. In addition, we consider a corpus $\T_p$ of trajectories, $(s_0, a_0, s_1, \hdots)$, relating to demonstrated policies. We assume the demonstrated policies are optimal w.r.t. to task MDPs sampled from $\D_R$. More specifically, trajectories in $\T_p$ are sampled from a distribution of permissible policies $\D_p$, defined by
$$
P_{\pi \sim \D_{p}}(\pi) = P(\pi \in \Pi^*_{\M_R})  ~\forall \pi,
$$
where $\Pi^*_{\M_R}$ is the set of optimal policies, which maximize the discounted return of $\M_R$.

\cmnt{
The constrained objective of Reinforcement Learning in the permissible policy setting is to find a policy that maximizes
\begin{align}
& \underset{\pi}{\text{max}} \nonumber
& & V^\pi(s) = \E^\pi \pth{\sum_{t=0}^\infty \gamma^t r(s_t,a_t) \given[\Big] s} \\
& \text{s.t.} \label{eq:opt_problem}
& & \pi \text{ is permissible}
\end{align}
}

We consider a state-action tuple $(s_t,a_t)$ at time $t$, and define its context of width $w$ as the sequence 
$c_w(s_t, a_t) = (s_{t-w}, a_{t-w}, \hdots s_{t-1}, a_{t-1}, s_{t+1}, a_{t+1}, \hdots s_{t+w}, a_{t+w})$.
We will sometimes refer to state-only contexts and action-only contexts as contexts containing only states or only actions, respectively. We denote the set of all possible contexts by $\C$. With abuse of notation, we write $(a,c)$ to denote the pair $<(s,a), c_w(s,a)>$.

In our work we will focus on the pointwise mutual information (PMI) of $(s,a)$ and its context $c$. Pointwise mutual information is an information-theoretic association measure between a pair of
discrete outcomes x and y, defined as
$
PMI(x,y) = \log \frac{P(x,y)}{P(x)P(y)}.
$
We will also consider the conditional $PMI$ denoted by $PMI(\cdot, \cdot | \cdot)$.

\subsection{Act2Vec}
We summarize the Skip-Gram neural embedding model introduced in \cite{skip_gram} and trained using the Negative-Sampling procedure  (SGNS) \cite{word2vec}. We use this procedure for representing actions (in contrast to words) using their contexts, and refer to this procedure as Act2Vec.

Every action $a \in \A$ is associated with a vector $\vec{a} \in \R^d$. In the same manner, every context $c \in \C$ is associated with a vector $\vec{c} \in \R^d$. In SGNS, we ask, does the pair $(a,c)$ come from $\D_p$? More specifically, we ask, what is the probability that $(a,c)$ came from $\D_p$? This probability, denoted by $P(\D_p = \text{true} ; a,c)$, is modeled as
$$
P(\D_p = \text{true} ; a,c) = \sigma(\vec{a}^T \vec{c}) = \frac{1}{1 + e^{-\vec{a}^T \vec{c}}}.
$$
Here, $\vec{a}$ and $\vec{c}$ are the model parameters to be learned. 

Negative Sampling \cite{word2vec} tries to minimize $P(\D_p = \text{false} ; a,c)$ for randomly sampled ``negative" examples, under the assumption that randomly selecting a context for a given action is likely to result in an unobserved $(a, c)$ pair. The local objective of every $(a,c)$ pair is thus given by
$$
\ell(a,c) = \log \sigma (\vec{a}^T \vec{c}) + k \E_{c_N \sim P_{\C}} \log \sigma(-\vec{a}^T \vec{c}_N),
$$
where $P_\C(c)$ \cmnt{$P_\C(c) = \frac{\sum_{c' \in \T_p} \indicator{c = c'} }{\abs{\T_p}}$} is the empirical unigram distribution of contexts in $\T_{p}$, and $k$ is the number of negative samples.

\cmnt{
The training procedure follows the majority of neural embedding models \cite{language_modeling}: (1) Scan the corpus (i.e., $\T_p$) and use $<$action, context$>$ pairs in the local window as training samples and (2) train the models to make actions useful for predicting contexts (or in reverse). }

The global objective can be written as a sum over losses of $(a,c)$ pairs in the corpus
$$
\ell = \sum_{a \in \A} \sum_{c \in \C}\#(a,c) \ell(a,c),
$$
where $\#(a,c)$ denotes the number of times the pair $(a, c)$ appears in $\T_p$.

\textbf{Relation to PMI:} Optimizing the global objective makes observed action-context pairs have similar embeddings, while scattering unobserved pairs. Intuitively, actions that appear in similar contexts should have similar embeddings. In fact, it was recently shown that SGNS implicitly factorizes the action-context matrix whose cells are the pointwise mutual information of the respective action and context pairs, shifted by a global constant \cite{PPMI_SGNS}. In what follows, we show that $PMI(a,c)$ is a useful measure for action representation in reinforcement learning.

\subsection{State-only context}

State-only contexts provide us with information about the environment as well as predictions of optimal trajectories. Let us consider the next-state context ${c = s'}$. More specifically, given a state action pair $(s,a)$ we are interested in the measure $PMI(a, s' | s)$, where here $s'$ is the random variable depicting the next state. The following lemma shows that when two actions $a_1, a_2$ have similar $PMI$ w.r.t. (with respect to) their next state context, they can be joined into a single action with a small change in value. 
\cmnt{
To see this, we start by defining a partitioning of the action space.
\begin{definition}
Let $\K = \braces{K_i}_{i=1}^K$, such that $K_i \subseteq \A$ and $\bigcup_i K_i = \A$. We say $\K$ a partitioning of $\A$.
\end{definition}
Next, we look at a special case in which we partition the action space so that actions are in the same partition if they have similar PMI w.r.t. $s'$.
\begin{definition}
We say the set $\K$ is an $\epsilon$ partitioning of $\A$ if
\begin{enumerate}
\item $\K$ is a partitioning of $\A$.
\item For every $K_i \in \K$, $a_1, a_2 \in K_i$, $s \in \s$ it holds that $\abs{PMI(a_1, s') - PMI(a_2, s')} < \epsilon$.
\end{enumerate}
\end{definition}
The next definition defines the policy that can't tell apart between actions that are in the same partiion.
\begin{definition}
Let $\pi \in \Pi$ be a stationary markovian policy, and let $\K$ be a partitioning of $\A$. Define the categorization policy of $\K$ given $\pi$ as follows
$$
\pi_\K(a|s) =
\begin{cases}
\frac{1}{\abs{K_i}}\sum_{j=1}^{\abs{K_i}} \pi(a_j^{(i)}|s) &,a \in K_i \\
\pi(a|s) &,o.w.
\end{cases}
$$
\end{definition}
The next theorem tells us that the categorization $\pi_\K$ has a similar value to $\pi$.
\begin{lemma}
\label{thm:state_context}
Let $\K$ be an $\epsilon$ partitioning of $\A$, $\pi \in \Pi$, and $\pi_\K$ be the categorization policy of $\pi$. Then
$$
\norm{V^\pi - V^{\pi_\K}}_\infty \leq \frac{6\gamma}{\pth{1-\gamma}^4}  R_{max} \sqrt{\epsilon}
$$
\end{lemma}
}

\begin{lemma}
\label{thm:state_context}
Let $K \subseteq \A$ s.t.
$$
\abs{PMI(a_1, s'|s) - PMI(a_2, s'| s)} < \epsilon, \forall a_1,a_2 \in K,
$$
where ${(a,s') \subset \tau \sim \pi \sim \D_p}, \forall a \in K$. \\
Let $\pi \sim \D_p$ and denote
$$
\pi_K(a|s) =
\begin{cases}
\frac{1}{\abs{K}}\sum_{j=1}^{\abs{K}}\pi(a_j|s) &,a \in K \\
\pi(a|s) &,o.w.
\end{cases}
$$
Then,
$$
\norm{V^\pi - V^{\pi_K}}_\infty \leq \frac{6\gamma}{\pth{1-\gamma}^4} \epsilon
$$
\end{lemma}

A proof to the lemma can be found in the supplementary material. The lemma illustrates that neglecting differences in actions of high proximity in embedding space has little effect on a policy's performance. While state contexts provide fruitful information, their PMI may be difficult to approximate in large state domains. For this reason, we turn to action-only contexts, as described next.

\cmnt{
\begin{proof}
$$
\abs{\log P(s' | s, a_1) - \log P(s' | s, a_2)} < \epsilon
$$
Then definitely it holds that
$$
\abs{P(s' | s, a_1) -  P(s' | s, a_2)} < \epsilon.
$$
By Pinsker's inequality
\begin{equation}
\label{eq:pinsker_inequality}
D_{TV} ( P || Q ) \leq \sqrt{ \frac{1}{2} D_{KL} (P || Q) }.
\end{equation}
It is also easy to show that for $\abs{a} < 1$
$$
\sum_{i=0}^\infty i a^i = a\frac{1 + 4a + a^2}{\pth{1-a}^4} \leq \frac{6a}{\pth{1-a}^4}.
$$
The rest of the proof follows from the simulation lemma in \href{http://ai.stanford.edu/~pabbeel/pubs/AbbeelNg_eaalirl_ICML2005long.pdf}{this paper}.
\end{proof}
}

\subsection{Action-only context}

Action-only contexts provide us with meaningful information when they are sampled from $\D_{p}$. To see this, consider the following property of action-only contexts. If action $a_1$ is more likely to be optimal than $a_2$, then any context that has a larger PMI with $a_1$ than $a_2$ will also be more likely to be optimal when chosen with $a_1$. Formally, let $\pi^* \sim \D_{p}$, $\tau^* \sim \pi^*$, and $a_1, a_2 \in \A, s \in \s$ such that

\begin{equation}
\label{eq:assumption}
P(\pi^*(a_1|s) \geq \pi^*(a_2|s)) \geq \frac{1}{2}.
\end{equation}
Let $c \in \C$. If 
\begin{equation}
\label{eq:pmi_assumption}
PMI(a_1, c|s) \geq PMI(a_2, c|s),
\end{equation} 
then
\begin{equation}
\label{eq:action_context_result}
P\pth{(s, a_1, c) \subseteq \tau^*} \geq P\pth{(s, a_2, c) \subseteq \tau^*}.
\end{equation} 

To show (\ref{eq:action_context_result}), we write the assumption in Equation (\ref{eq:pmi_assumption}) explicitly:
$$
\frac{P(a_1, c | s)}{P(a_1 | s)P(c | s)} \geq \frac{P(a_2, c | s)}{P(a_2 | s)P(c | s)}.
$$
Next, due to the definition of $\D_{p}$, we have that
$$
\frac{P((s, a_1, c) \subseteq \tau^*)}{P(a_1 | s)P(c | s)} \geq \frac{P((s, a_2, c) \subseteq \tau^*)}{P(a_2 | s)P(c | s)} .
$$
Giving us
\begin{align*}
P((s, a_1, c) \subseteq \tau^*) 
&\geq \frac{P(a_1|s)}{P(a_2|s)} P((s,a_2, c) \subseteq \tau^*) \\
&\geq P((s,a_2, c) \subseteq \tau^*),
\end{align*}

where the last step is due to our assumption in \mbox{Equation \ref{eq:assumption}.}

In most practical settings, contexts based embeddings that use state contexts are difficult to train using SGNS. In contrast, action-only contexts usually consist of orders of magnitude less elements. For this reason, in Section \ref{sec:empirical} we experiment with action-only contexts, showing that semantics can be learned even when states are ignored.

\section{Act2Vec for function approximation}
\label{sec:func_approx}

\begin{figure*}[ht!]
\vskip 0.1in
\begin{center}
\begin{subfigure}{0.3\textwidth}
\includegraphics[width=\linewidth]{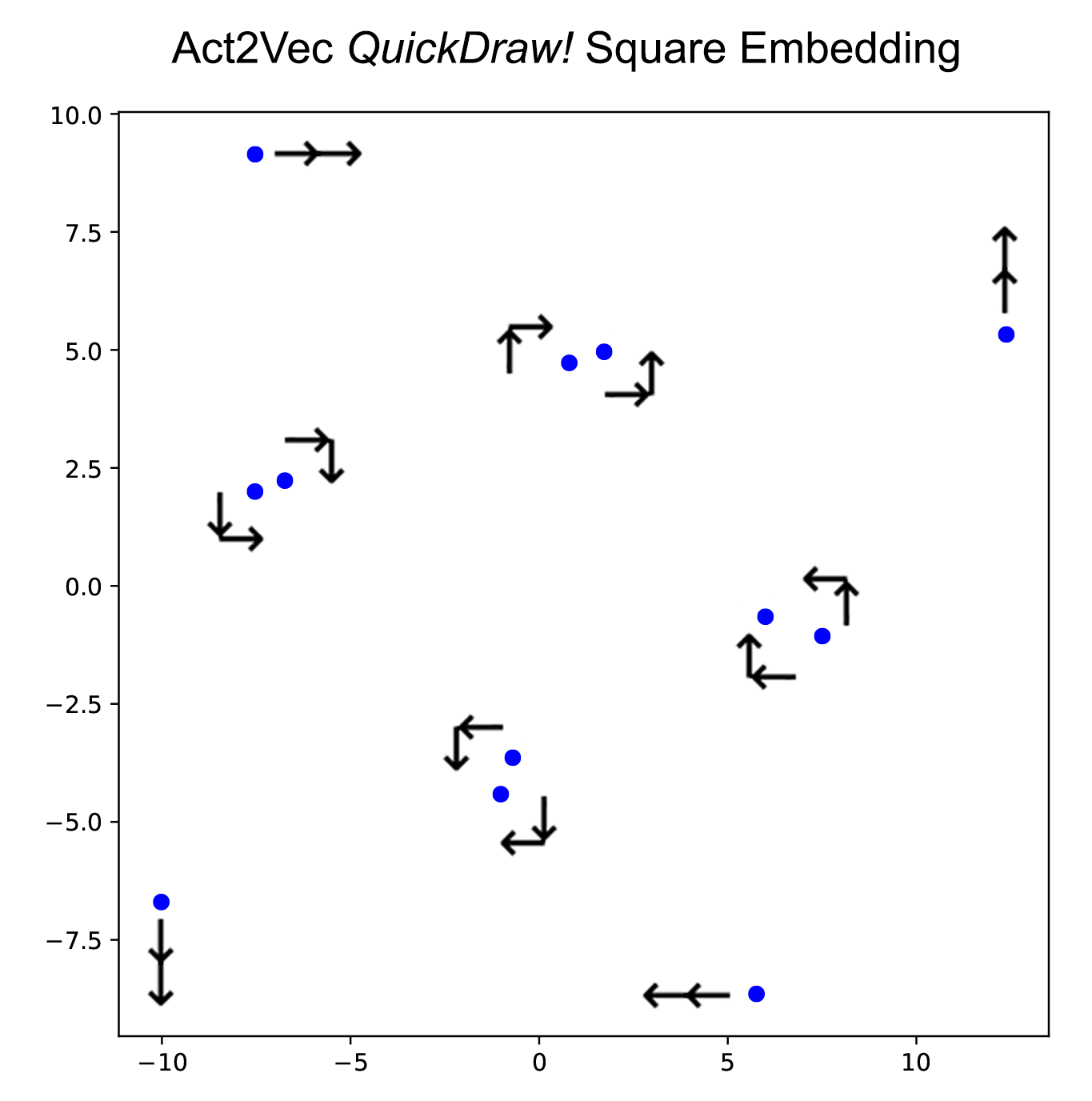}
\caption{}
\end{subfigure}
\begin{subfigure}{0.33\textwidth}
\includegraphics[width=\linewidth]{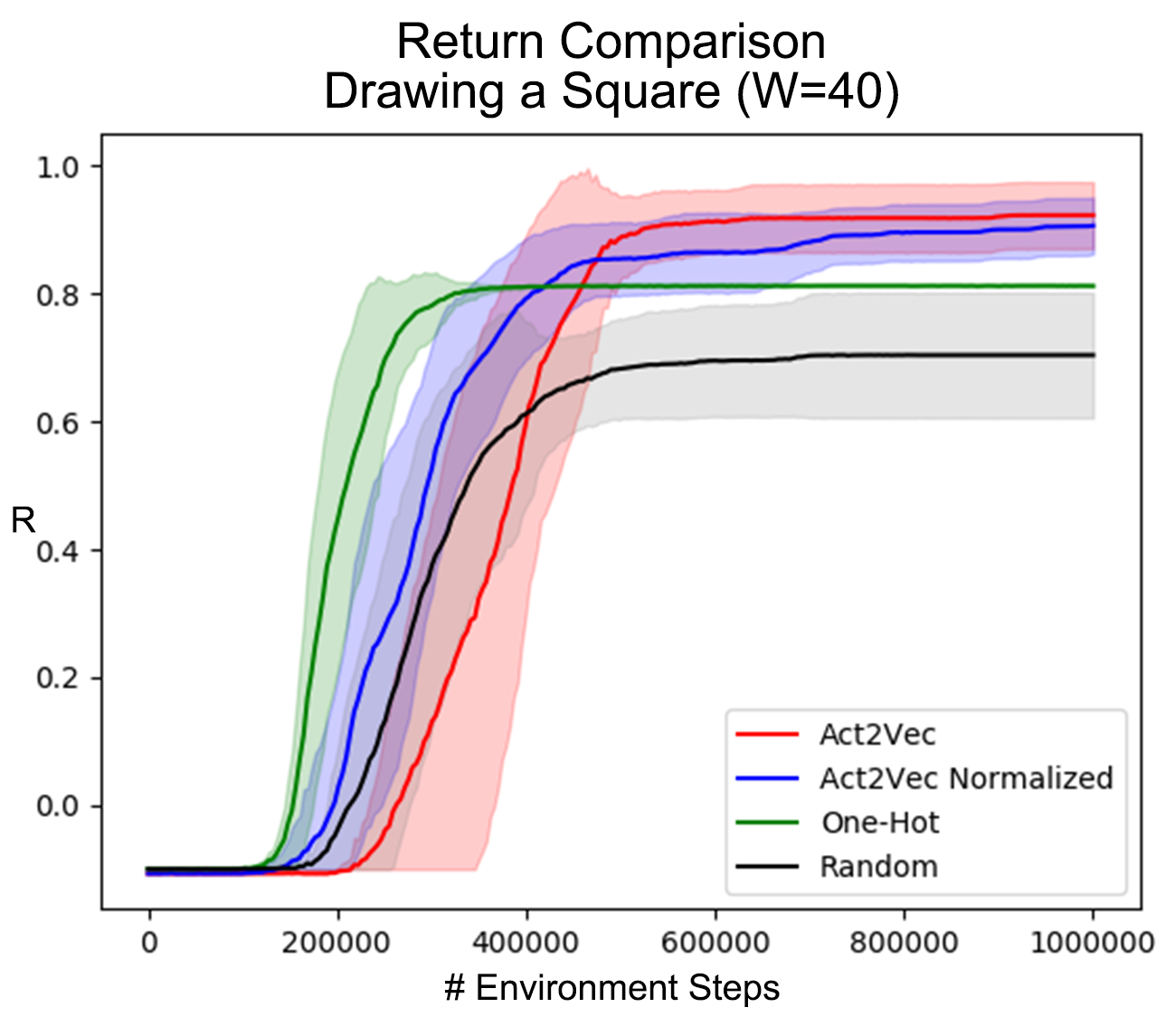}
\caption{}
\end{subfigure}
\begin{subfigure}{0.33\textwidth}
\includegraphics[width=\linewidth]{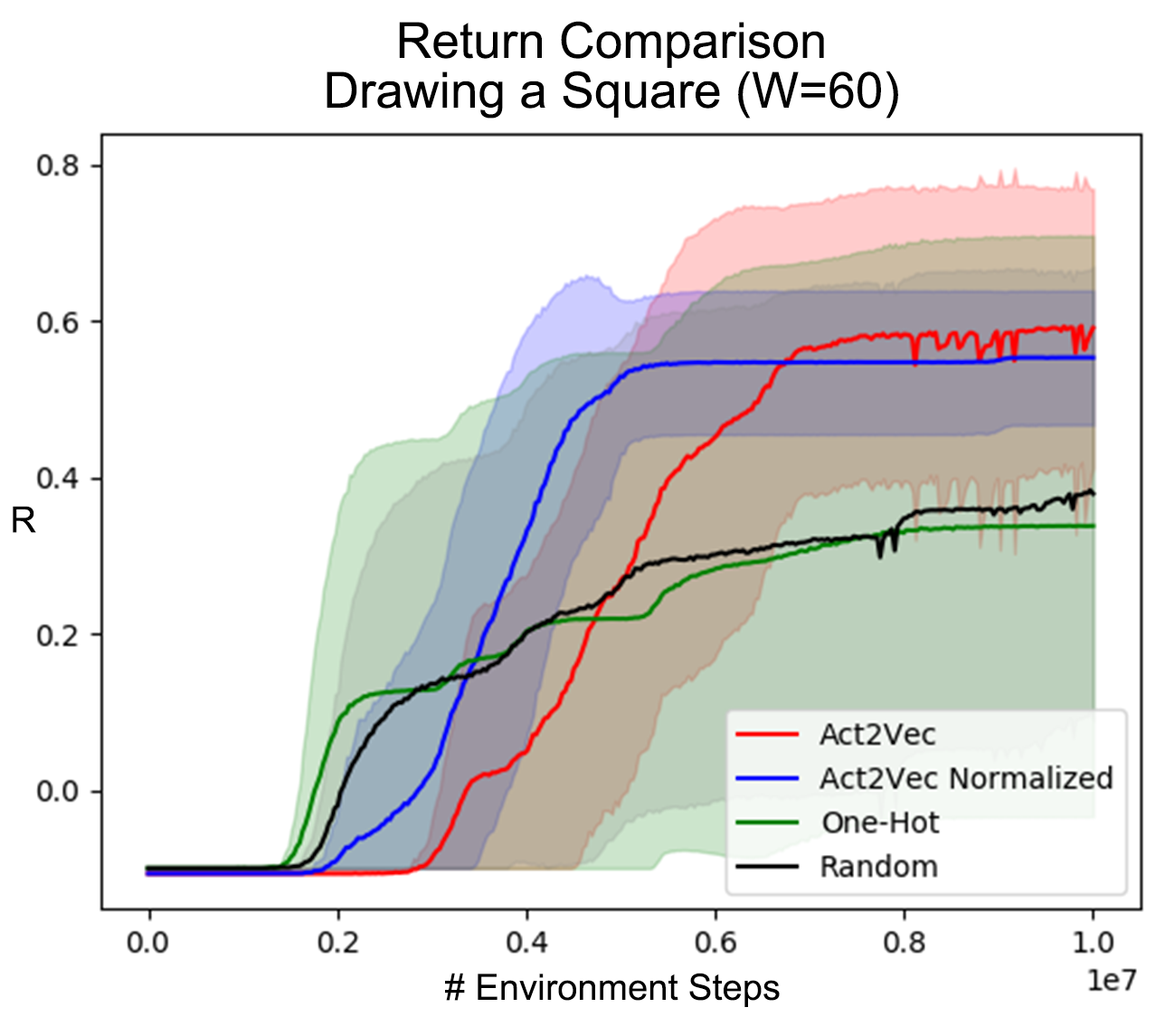}
\caption{}
\end{subfigure}
\vskip -0.1in
\caption{\textbf{(a):} Plot shows Act2Vec embedding of \textit{QuickDraw!}'s square category strokes. Actions are positioned according to direction. Corner strokes are organized in relation to specific directions, with symmetry w.r.t. direction in which squares were drawn. \\
\textbf{(b-c):} Comparison of state representations for drawing squares with different edge lengths ($W=40, 60$) . The state was represented as the sum of previous action embeddings. Results show superiority of Act2Vec embedding as opposed to one-hot and random embeddings. }
\label{fig:quickdraw}
\end{center}
\vskip -0.2in
\end{figure*}

Lemma \ref{thm:state_context} gives us intuition as to why actions with similar contexts are in essence of similar importance to overall performance. We use this insight to construct algorithms that depend on similarity between actions in the latent space. Here, we consider the set of discrete \mbox{actions $\A$}. Equivalently, as we will see in Section \ref{sec:navigation}, this set of actions can be augmented to the set of all action sequences $\{(a_1, \hdots, a_k)\}$. We are thus concerned with approximating the $Q$-value of state-action pairs $Q(s,a)$.

\textbf{$Q$-Embedding:} When implemented using neural networks, $Q$-Learning with function approximation consists of approximating the $Q$-value of state-action pairs by
\begin{equation}
\label{eq:standard_func_approx}
\hat{Q}(s,a) = w_a^T\phi(s),
\end{equation}
where $\phi(s)$ are the features learned by the network, and $w_a$ are the linear weights learned in the final layer of the network. When the number of actions is large, this process becomes impractical. In NLP domains, it was recently suggested to use word embeddings to approximate the $Q$-function \cite{nlp_rl}  as
\begin{equation}
\label{eq:embd_func_approx}
\hat{Q}(s,a) = \psi(a)^T \phi(s),
\end{equation}
where $\psi(a)$ are the learned features extracted from the embedding of a word $a$. Similar words become close in embedding space, thereby outputting similar $Q$-Values. This approach can also be applied on action embeddings trained using Act2Vec. These action representations adhere inherent similarities, allowing one to approximate their $Q$-values, while effectively obtaining complexity of smaller dimension. We will refer to the approximation in \mbox{Equation \ref{eq:embd_func_approx}} as $Q$-Embedding.

\textbf{$k$-Exp:} In $Q$-Learning, the most fundamental exploration strategy consists of uniformly sampling an action. When the space of actions is large, this process becomes infeasible. In these cases, action representations can be leveraged to construct improved exploration strategies. We introduce a new method of exploration based on action embeddings, which we call $k$-Exp. $k$-Exp is a straightforward extension of uniform sampling. First, the action embedding space is divided into $k$ clusters using a clustering algorithm (e.g., $k$-means). The exploration process then follows two steps: (1) Sample a cluster uniformly, and (2) given a cluster, uniformly sample an action within it. $k$-Exp ensures actions that have semantically different meanings are sampled uniformly, thereby improving approximation of $Q$-values.

In Section \ref{sec:navigation} we compare $Q$-Embedding with \mbox{$k$-Exp} to $Q$-learning with uniform exploration, demonstrating the advantage of using action representations.

\section{The Semantics of Actions}
\label{sec:empirical}

Word embeddings have shown to capture large numbers of syntactic and semantic word relationships \cite{word2vec}. Motivated by this, as well as their relation to $PMI$, we demonstrate similar interpretation on several reinforcement learning environments. This section is divided into three parts. The first and second parts of this section consider the tasks of drawing a square and navigating in 3d space. In both parts, we demonstrate the semantics captured by actions in their respective domains. We demonstrate the effectiveness of Act2Vec in representing a state using the sequence of previously taken actions (see Figure \ref{fig:rl_diagram}). We then demonstrate the use of Act2Vec with $Q$-Embedding and $k$-Exp. Finally, in the third part of this section we demonstrate the semantic nature of actions learned in the complex strategy game of StarCraft II. 

\subsection{Drawing}
\label{sec:drawing}
 \cmnt{
Learning the process of drawing has been extensively studied in recent years \cite{sketch-rnn, generative1, generative2, gan_painting}. Recently, the sketch-rnn model \cite{sketch-rnn} used sequence-to-sequence learning \cite{seq2seq} to produce impressive sketches both unconditionally and conditioned on data. Their framework was capable of producing sketches of common objects in a vector format. The sketch-rnn model implicitly learns to embed strokes (i.e., actions) by their use in the reconstruction procedure of an image. While this latent space holds valuable information, its creation process is dependent on the ability of generating images from the given distribution. On the contrary, Act2Vec only relies on the prediction quality of short strokes from their context \footnote{Context-based embeddings can be used by a learning model to efficiently learn task-specific embeddings.}, making it independent of the task, algorithm, or model of choice.  }

We undertook the task of teaching an agent to draw a square given a sparse reward signal. The action space consisted of 12 types of strokes: Left, Right, Up, Down, and all combinations of corners (e.g., Left+Up). The sparse reward provided the agent with feedback only once she had completed her drawing, with positive feedback only when the drawn shape was rectangular. Technical details of the environment can be found in the supplementary material. \cmnt{ The reward function was given by ${R = \frac{\sum_{i=1}^4 \min \pth{W, l_i}}{4W}}$, where $l_i$ denotes the length of the $i^{th}$ side of the drawn shape, and $W$ is the desired length of each side. When the drawn shape had more or less than four sides, a reward of $-0.1$ was given.} We trained Act2Vec with action-only context \cmnt{,embedding dimension $d=10$ and window $w=2$} over a corpus of 70,000 human-made drawings in the ``square category" of the \textit{Quick, Draw!} \cite{quickdraw} dataset. Projections of these embeddings are depicted in Figure \ref{fig:quickdraw}(a). The embedding space projection reflects our knowledge of these action strokes. The space is divided into 4 general regions, consisting of strokes in each of the main axis directions. Strokes relating to corners of the squares are centered in distinct clusters, each in proximity to an appropriate direction. The embedding space presents evident symmetry w.r.t. clockwise vs. counterclockwise drawings of squares. 

The action space in this environment is relatively small. One way of representing the state is through the set of all previous actions, since in this case $s_t = \cup_{i=0}^{t-1} a_i$. The state was therefore represented as the vector equal to the sum of previous action vectors. We compared three types of action embeddings for representing states: Act2Vec, normalized Act2Vec (using the $l_2$ norm), one-hot, and randomized embeddings. Figure \ref{fig:quickdraw}(b,c) shows results of these representations for different square sizes. Act2Vec proved to be superior on all tasks, especially with increased horizon - where the sparse reward signal drastically affected performance. We also note that normalized Act2Vec achieved similar results with higher efficiency. In addition, all methods but Act2Vec had high variance in their performance over trials, implying they were dependent on the network's initialization. A detailed overview of the training process can be found in the supplementary material.

\subsection{Navigation}
\label{sec:navigation}

\begin{figure*}[ht!]
\vskip 0.2in
\begin{center}
\begin{subfigure}{0.31\textwidth}
\includegraphics[width=\linewidth]{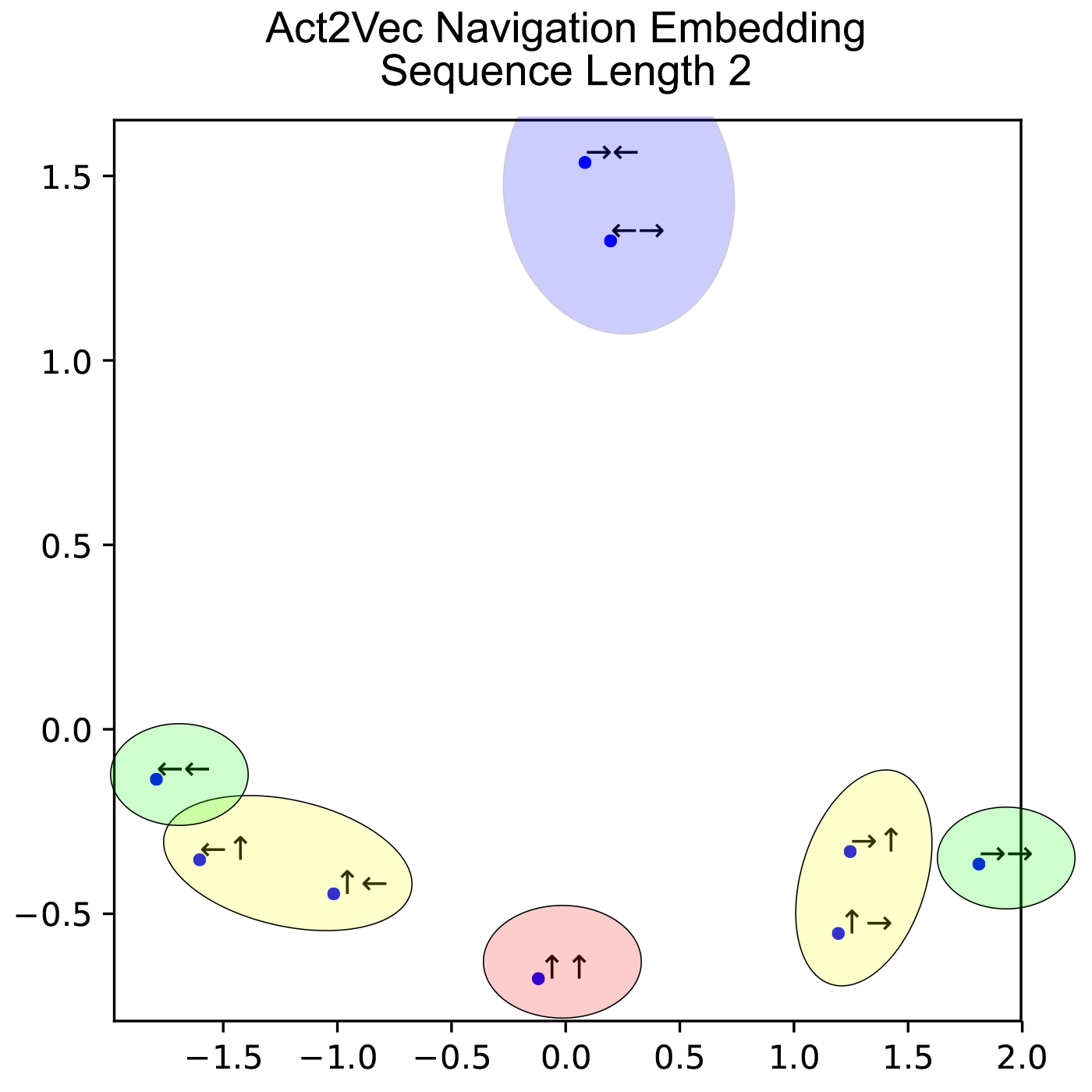}
\caption{}\label{fig:1a}
\end{subfigure}
\begin{subfigure}{0.31\textwidth}
\includegraphics[width=\linewidth]{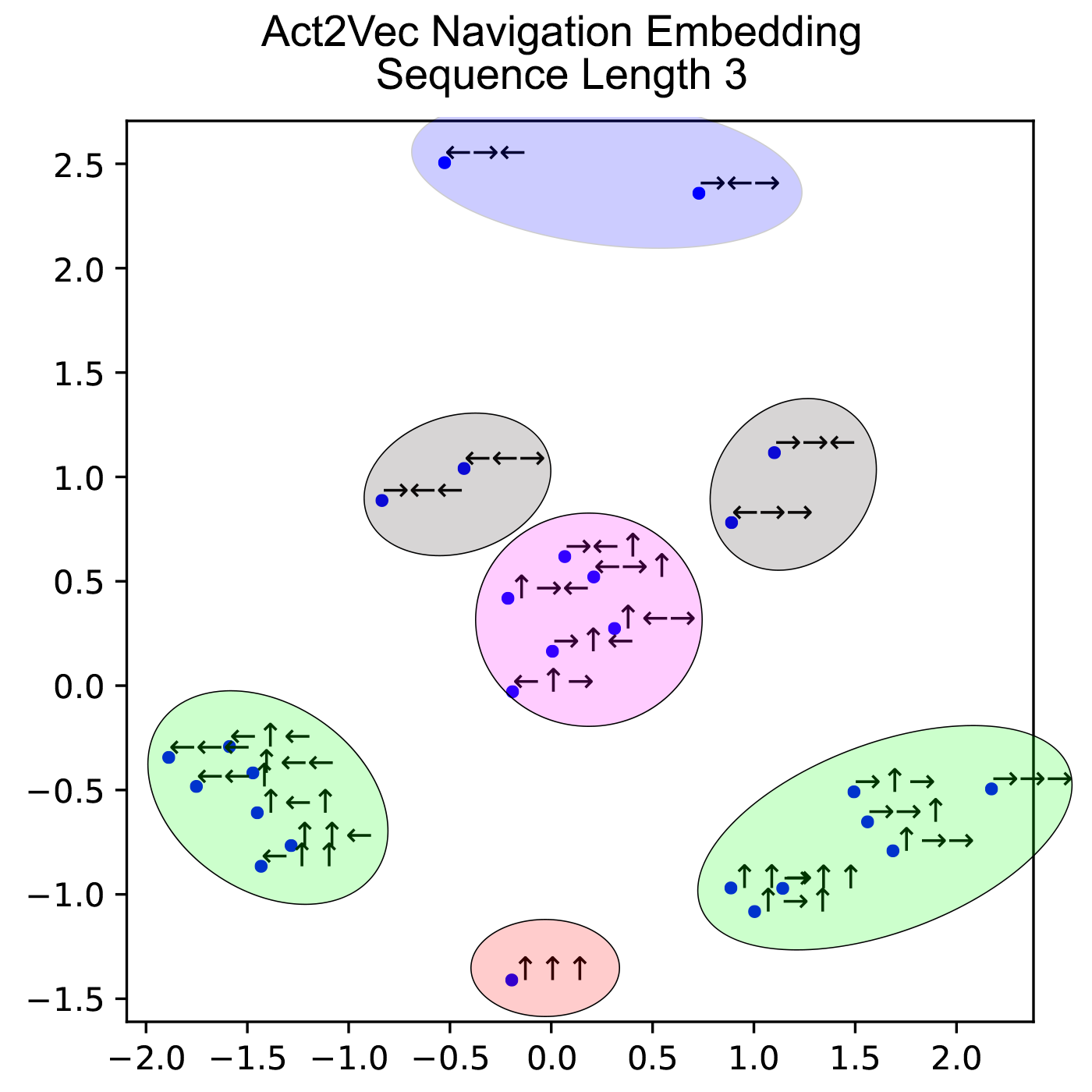}
\caption{}\label{fig:2a}
\end{subfigure}
\begin{subfigure}{0.37\textwidth}
\includegraphics[width=\linewidth]{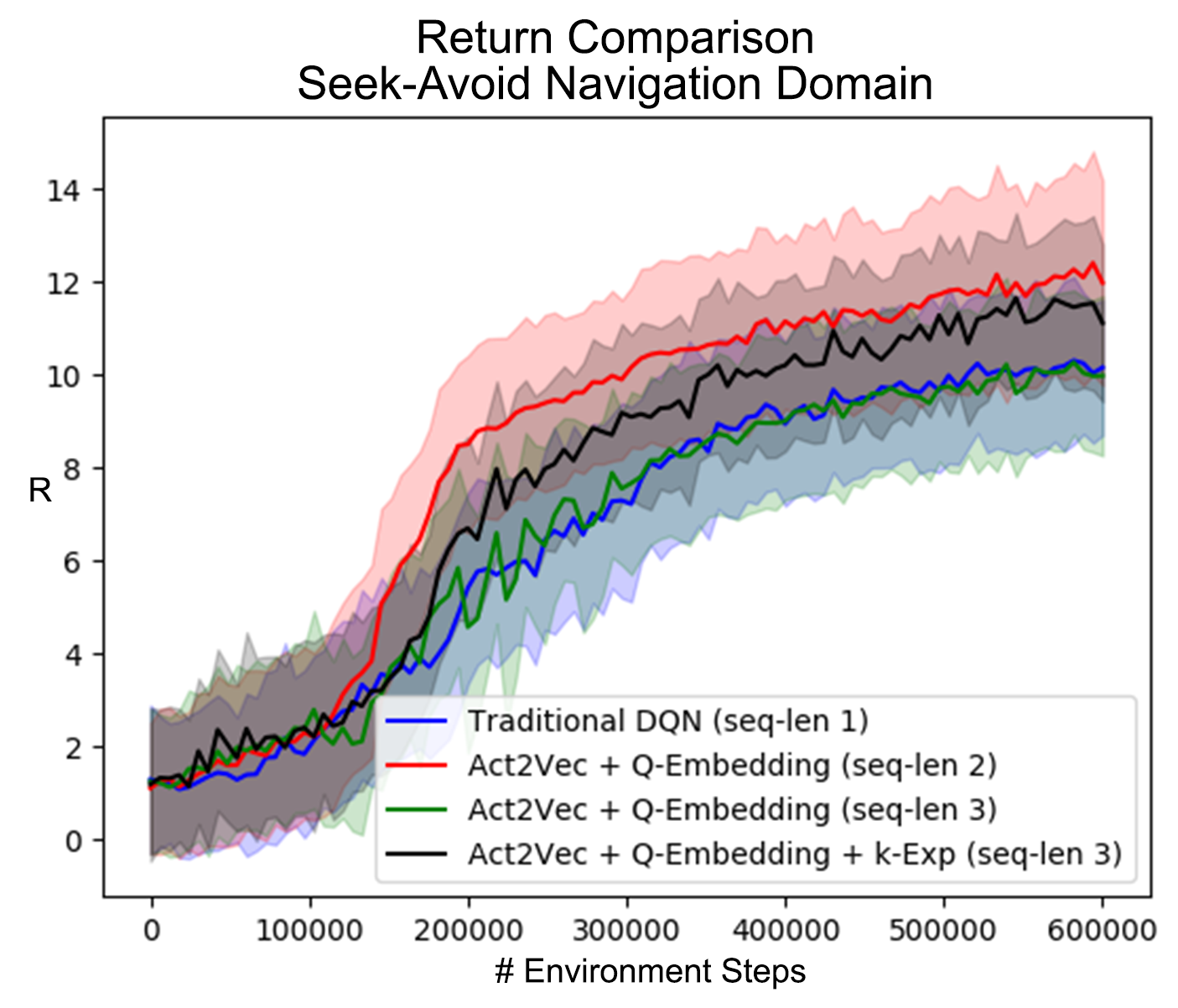}
\caption{}\label{fig:3a}
\end{subfigure}
\vskip -0.1in
\caption{\textbf{(a-b):} The action corpus of size 3000 actions was generated by actions executed in a 2d navigation domain consisting of three actions: move forward ($\uparrow$), \mbox{rotate view left ($\leftarrow$)}, and \mbox{rotate view right ($\rightarrow$)}. Plots show PCA projection of Act2Vec embedding for sequences of length 2 and 3. \cmnt{Plots show distinguishable clusters as well as global and local symmetry w.r.t. action sequences. As an example, examining the left plot of sequences of length 2, we notice symmetry w.r.t. sequences containing the $\leftarrow$ and the $\rightarrow$ actions. The fact that the sequences $(\leftarrow, \rightarrow)$ and $(\rightarrow, \leftarrow)$ are equivalent and distinctive to all other sequences is represented through a distant cluster. Action sequences closer to the bottom-center part of the plot can be related to a forward momentum in space.} 
\textbf{(c):} Comparison of techniques on the Seek-Avoid environment. Plots show results for different sequence lengths, with and without $Q$-Embedding. Sequences of length $3$ only showed improvement when cluster based exploration was used.   }
\label{fig:navig1}
\end{center}
\vskip -0.2in
\end{figure*}

In this section we demonstrate how sequences of actions can be embedded using Act2Vec. We then show how embeddings based on trajectories captured in a simple navigation domain can \textit{transfer knowledge} to a more complex navigation domain, thus improving its learning efficiency.

In physical domains, acceptable movements of objects in space are frequently characterized by smoothness of motion. As such, when we open a door, we move our hand smoothly through space until reaching the knob, after which we complete a smooth rotation of the doorknob. An agent learning in such an environment may tend to explore in a manner that does not adhere to patterns of such permissible behavior (e.g., by uniformly choosing an arbitrary action). Moreover, when inspecting individual tasks, actions incorporate various properties that are particular to the task at hand. Looking left and then right may be a useless sequence of actions when the objective is to reach a goal in space, while essential for tasks where information gathered by looking left contributes to the overall knowledge of an objective on the right (e.g., looking around the walls of a room). In both cases, looking left and immediately right is without question distinct to looking left and then looking left again. These semantics, when captured properly, can assist in solving any navigation task. 

When studying the task of navigation, one is free to determine an action space of choice. In most applications, the primitive space of actions is either defined by fixed increments of movement and rotation or by physical forces. Let us consider the former case and more specifically examine the action space consisting of moving forward (marked by $\uparrow$), and rotating our view to the left or to the right (marked by $\leftarrow$ and $\rightarrow$, respectively). These three actions are in essence sufficient for solving most navigation tasks. Nonetheless, semantic relations w.r.t. these actions become particularly evident when action sequences are used. For this case, we study the augmented action space consisting of all action sequences of length $k$, i.e., ${(a_1, a_2, \hdots, a_k) \in \A^k}$. For example, for the case of $k=2$, the action sequence $(\uparrow, \uparrow)$ would relate to taking the \mbox{action $\uparrow$} twice, thereby moving two units forward in the world, whereas $(\uparrow,  \leftarrow)$ would relate to moving one unit forward in the world and then rotating our view one unit to the left. This augmented action space holds interesting features, as we see next. 

We trained Act2Vec with action-only context on a corpus of 3000 actions taken from a 2d navigation domain consisting of randomly generated walls. Given a random goal location, we captured actions played by a human player in reaching the goal. Figure \ref{fig:navig1} (a,b) shows Principal Component Analysis (PCA) projections of the resulting embeddings for action sequences of length $k=2,3$. Examining the resulting space, we find two interesting phenomena. First, the embedding space is divided into several logical clusters, each relating to an aspect of movement. In these, sequences are divided according to their forward momentum as well as direction. Second, we observe symmetry w.r.t. the vertical axis, relating to looking left and right. These symmetrical relations capture our understanding of the consequences of executing these action sequences. In the next part of this section we use these learned embeddings in a \textit{different} navigation domain, teaching an agent how to navigate while avoiding unwanted objects.

\begin{figure*}[ht!]
\vskip 0.2in
\begin{center}
\begin{subfigure}{0.41\textwidth}
\includegraphics[width=\linewidth]{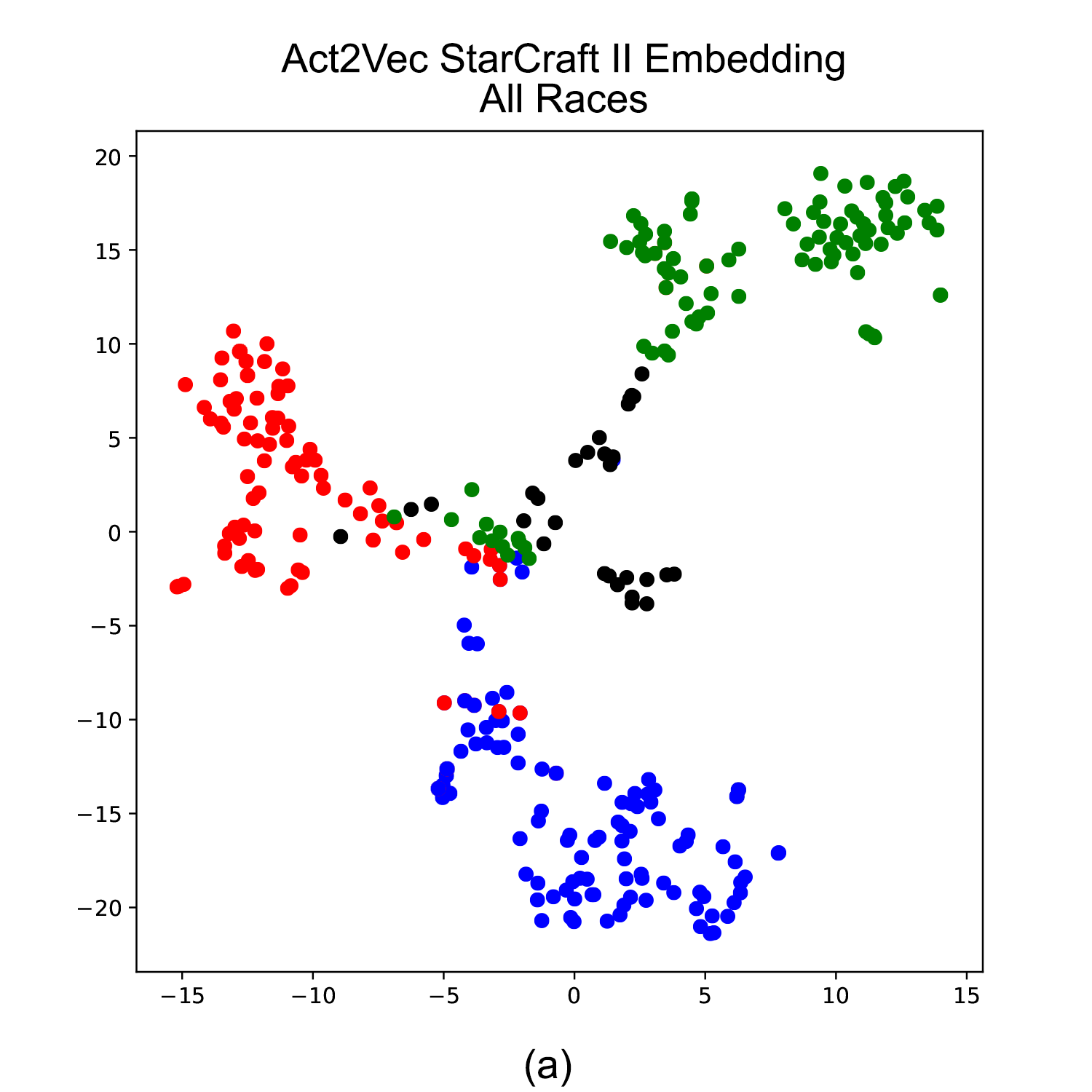}
\end{subfigure}
\begin{subfigure}{0.55\textwidth}
\includegraphics[width=\linewidth]{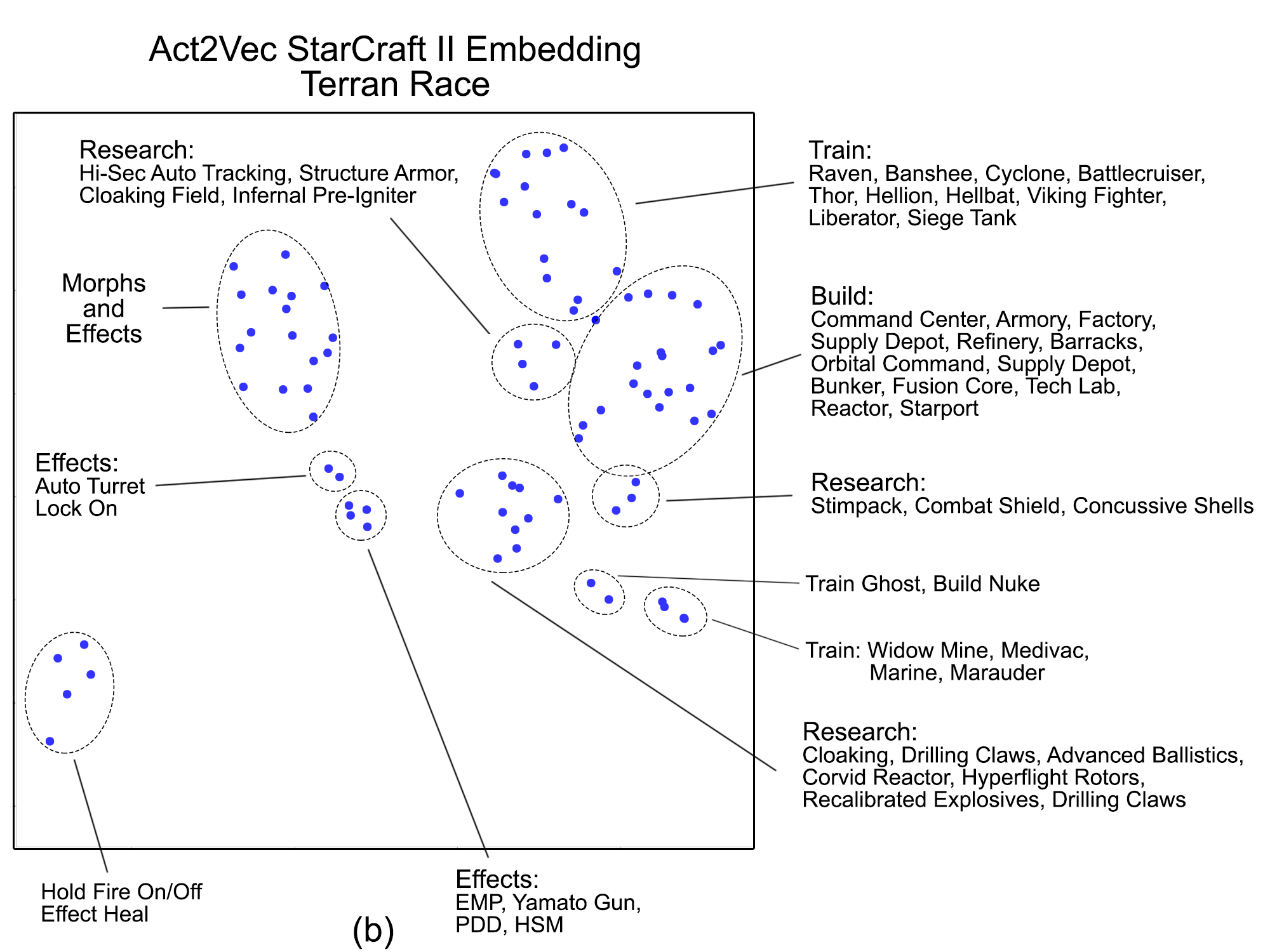}
\end{subfigure}
\caption{ Plots show t-SNE embedding of Starcraft II action functions for all races (a) as well as the Terran race (b). Action representation distinguish between all three races through separate clusters. Plot (b) depicts clusters based on categorial types: building, training, researching, and effects. Clusters based on common player strategies appear in various parts of the embedding space.  }
\label{fig:starcraft}
\end{center}
\vskip -0.2in
\end{figure*}

\textbf{Knowledge Transfer:} We tested the Act2Vec vector embedding trained on the 2d navigation domain on the DeepMind Lab \cite{deepmind_lab} Seek-Avoid environment \footnote{While the tasks of navigation in the 2d domain and the Seek-Avoid domain are different, the usage of their actions presents similar semantic knowledge, thereby incorporating transfer of knowledge between these domains.}. Here, an agent must navigate in 3d space, learning to collect good apples while avoiding bad ones. The agent was trained using Q-learning with function approximation. We tested sequences of length $k=1,2,3$ using both methods of function approximation \mbox{(Equations \ref{eq:standard_func_approx} and \ref{eq:embd_func_approx}).} 

Results, as depicted in \mbox{Figure \ref{fig:navig1}(c)}, show superiority of using embeddings to approximate Q-values. Action sequences of length $k=2$ showed superiority over $k=1$ with \mbox{$\epsilon$-uniform} exploration, with $20$ percent increase in total reward. Sequences of length $3$ did not exhibit an increase in performance. We speculate this is due to the uniform exploration process. To overcome this matter, we used k-means in order to find clusters in the action embedding space. We then evaluated $k$-Exp on the resulting clusters. While results were dependent on the initialization of clusters (due to randomization of $k$-means), sequences of length $3$ showed increase in total reward. Additional technical details can be found in the supplementary material.

\textbf{Regularization:} Learning to navigate using sequence of actions led to smoothness of motion in the resulting tests. This property arose due to the unique clustering of sequences. Particularly, the sequences $(\leftarrow, \rightarrow)$ and ${(\rightarrow, \leftarrow)}$ were discarded by the agent, allowing for a smooth flow in navigation. Consequently, this property indicates that action sequences act as a regularizer to the navigation task.

\subsection{StarCraft II}
\label{sec:starcraft}

StarCraft II is a popular video game that presents a very hard challenge for reinforcement learning. Its main difficulties include a huge state and action space as well as a long-time horizon of thousands of states. The consequences of any single action (in particular, early decisions) are typically only observed many frames later, posing difficulties in temporal credit assignment and exploration. Act2Vec offers an opportunity to mitigate some of these challenges, by finding reliable similarities and relations in the action space. 

We used a corpus of over a million game replays played by professional and amateur players. The corpus contained over 2 billions played actions, which on average are equivalent to 100 years of consecutive gameplay. The action space was represented by over 500 action functions, each with 13 types of possible arguments, as described in \cite{starcraft}. We trained Act2Vec with action-only context to embed the action functions into action vectors of dimension $d$, ignoring any action arguments. T-SNE \cite{tsne} projections of the resulting action embeddings are depicted in Figure \ref{fig:starcraft}. 

In StarCraft II players choose to play one of three species: Terran, Protoss, or Zerg. Once a player chooses her race, she must defeat her opponent through strategic construction of buildings, training of units, movement of units, research of abilities, and more. While a myriad strategies exist, expert players operate in conventional forms. Each race admits to different strategies due to its unique buildings, units, and abilities. Embeddings depicted in Figure \ref{fig:starcraft}(a) show distinct clusters of the three different races. Moreover, actions that are commonly used by all three races are projected to a central position with equal distance to all race clusters. Figure \ref{fig:starcraft}(b) details a separate t-SNE projection of the Terran race action space. Embeddings are clustered into regions with definite, distinct types, including: training of units, construction of buildings, research, and effects. While these actions seem arbitrary in their raw form, Act2Vec, through context-based embedding, captures their relations through meaningful clusters. 

Careful scrutiny of Figure \ref{fig:starcraft}(b), shows interesting captured semantics, which reveal common game strategies. As an example, let us analyze the cluster containing actions relating to training Marines, Marauders, Medivacs, and WidowMines. Marines are an all-purpose infantry unit, while Marauders being almost opposite to the Marine units, are effective at the front of an engagement to take damage for Marines. Medivacs are dual purpose dropships and healers. They are common in combination with Marines and Marauders, as they can drop numbers of Marines and Marauders and then support them with healing. A Widow Mine is a light mechanical mine that deals damage to ground or air units. Widow Mines are used, as a standard opening move, in conjunction with Marines and Medivacs, to zone out opposing mineral fields. Other examples of strategic clusters include the Ghost and the Nuke, which are commonly used together, as well as the Stimpack, Combat Shield, and Concussive Shells abilities, which are Marine and Marauder upgrades, all researched at the Tech Lab attached to a Barracks.

The semantic representation of actions in StarCraft II illustrates how low dimensional information can be extracted from high dimensional data through an elementary process. It further emphasizes that knowledge implicitly incorporated in Act2Vec embeddings can be compactly represented without the need to solve the (many times challenging) task at hand.

\section{Related Work}
\label{sec:related_work}

\textbf{Action Embedding:} \cite{dulac2015deep} proposed to embed discrete actions into a continuous space. They then find optimal actions using a nearest neighbor approach. They do not, however, offer a method in which such action representations can be found. Most related to our work is that of \cite{chandak2019learning} in which an embedding is used \textit{as part of the policy's structure} in order to train an agent. Our work provides a complementary aspect, with an approach to directly inject prior knowledge from expert data. In addition, we are able to capture semantics without the need to solve the task at hand. 

\textbf{Representation Learning:} Representation learning is concerned with finding an appropriate representation of data in order to perform a machine learning task \cite{representation}. In particular, deep learning exploits this concept by its very nature \cite{dqn}. Other work related to representation in RL include Predictive State Representations (PSR) \cite{psr}, which capture the state as a vector of predictions of future outcomes, and a Heuristic Embedding of Markov Processes (HEMP) \cite{hemp}, which learns to embed transition probabilities using an energy-based optimization problem. In contrast to these, actions are less likely to be affected by the ``curse of dimensionality" that is inherent in states. 

One of the most fundamental work in the field of NLP is word embedding \cite{word2vec, glove, ngram2vec}, where low-dimensional word representations are learned from unlabeled corpora. Among most word embedding models, Word2Vec \cite{word2vec} (trained using SGNS) gains its popularity due to its effectiveness and efficiency. It achieves state-of-the-art performance on a range of linguistic tasks within a fraction of the time needed by previous techniques. In a similar fashion, Act2Vec represents actions by their context. It is able to capture meaningful relations between actions - used to improve RL agents in a variety of tasks.

\textbf{Learning from Demonstration (LfD):} 
Imitation learning is primarily concerned with matching the performance of a demonstrator \cite{imitation2, imitation1, imitation3}. Demonstrations typically consist of sequences of state-action pairs $\{(s_0, a_0), \hdots, (s_n, a_n)\}$, from which an agent must derive a policy that reproduces and generalizes the demonstrations. While we train Act2Vec using a similar corpus, we do not attempt to generalize the demonstrator's mapping $\s \to \A$. Our key claim is that vital information is present in the order in which actions are taken. More specifically, the context of an action masks acceptable forms and manners of usage. These natural semantics cannot be generalized from state-to-action mappings, and can be difficult for reinforcement learning agents to capture. By using finite action contexts we are able to create meaningful representations that capture relations and similarities between actions.

\textbf{Multi-Task RL:}
Multitask learning learns related tasks with a shared representation in parallel, leveraging information in related tasks as an inductive bias, to improve generalization, and to help improve learning for all tasks \cite{multi-task1, multi-task2, transfer1, transfer2, transfer3}. In our setting, actions are represented using trajectories sampled from permissible policies\footnote{In our setting, policies are given as optimal solutions to tasks. In practical settings, due to finite context widths, policies need not be optimal in order to capture relevant, meaningful semantics.}. These representations advise on correct operations for learning new tasks, though they only incorporate local, relational information. They provide an approach to implicitly incorporate prior knowledge through representation. Act2Vec can thus be used to improve efficiency of multi-task RL methods.


\textbf{Skill Embedding:}
Concisely representing skills allow for efficient reuse when learning complex tasks \cite{skill-embedding1, skill-embedding2, skill-embedding3}. Many methods use latent variables and entropy constraints to decrease the uncertainty of identifying an option, allowing for more versatile solutions \cite{skill-entropy1, skill-entropy2, skill-entropy3}. While these latent representations enhance efficiency, their creation process is dependent on the agent's ability to solve the task at hand. The benefit of using data generated by human demonstrations is that it lets one learn expressive representations without the need to solve any task. Moreover, much of the knowledge that is implicitly acquired from human trajectories may be unattainable by an RL agent. As an example of such a scenario we depict action embeddings learned from human replays in StarCraft II (see Section \ref{sec:starcraft}, Figure \ref{fig:starcraft}). While up-to-date RL algorithms have yet to overcome the obstacles and challenges in such problems, Act2Vec efficiently captures evident, valuable relations between actions.

\section{Discussion and Future Work}
\label{sec:discussion}

If we recognize actions as symbols of a natural language, and regard this language as ``expressive", we imply that it is an instrument of inner mental states. Even by careful introspection, we know little about these hidden mental states, but by regarding actions as thoughts, beliefs, strategies, we limit our inquiry to what is objective. We therefore describe actions as modes of behavior in relation to the other elements in the context of situation.

\cmnt{
In many RL problems, the input data is high-dimensional and abundant, and reward feedback is scarce or absent. In such situations, finding the right representation of the data can be the key to solving the problem. Convolutional and recurrent neural models have been shown to learn low dimensional features from high-dimensional data, making it more amenable to learning. These models work well on input data that has strong spatial properties. While a substantial portion of current RL benchmarks rely on visual imagery, the input data of frequent problems carries information that does not hold distinct spatial properties. Action representation generated by Act2Vec admit spatial properties due to their PMI relations, as they were generated in expressive domains which encode relations through their contexts. }


When provided with a structured embedding space, one can efficiently eliminate actions. When the number of actions is large, a substantial portion of actions can be eliminated due to their proximity to known sub-optimal actions. When the number of actions is substantially small, action sequences can be encoded instead, establishing an extended action space in which similar elimination techniques can be applied. 

Recognizing elements of the input as segments of an expressive language allow us to create representations that adhere unique structural characteristics. While the scope of this paper focused on action representation, distributional embedding techniques may also be used to efficiently represent states, policies, or rewards through appropriate contexts. Interpreting these elements relative to their context withholds an array of possibilities for future research.

Lastly, we note that distributional representations can be useful for debugging RL algorithms and their learning process. By visualizing trajectories in the action embedding space, an overseer is able to supervise over an agents progress, finding flaws and improving learning efficiency.

\bibliography{bibfile}
\bibliographystyle{icml2019}

\onecolumn

\section{Supplementary Material}

\subsection{Proof of Lemma \ref{thm:state_context}}

We show a more general formulation of the lemma. More specifically, we show that it holds for
$$
\pi_K(a|s) =
\begin{cases}
\mu(a|s)\frac{\sum_{a' \in K} \pi(a'|s)}{\sum_{a' \in K} \mu(a'|s)} &,a \in K \\
\pi(a|s) &,o.w.
\end{cases},
$$
where $\mu$ is any (possibly adversarial) distribution over the set of actions in $K$. In the case of the lemma, $\mu$ is the uniform distribution.

Writing the assumption explicitly we have that for all $a_1, a_2 \in K$
$$
\abs{\frac{P_{(a_1, s') \sim \D_p}(a_1, s'|s)}{P_{a_1 \sim \D_p}(a_1|s)P_{s' \sim \D_p}(s'|s)} - \frac{P_{(a_2, s') \sim \D_p}(a_2, s'|s)}{P_{a_2 \sim \D_p}(a_2|s)P_{s' \sim \D_p}(s'|s)}} < \epsilon,
$$
which can be rewritten
$$
\abs{\frac{P(s'|s,a_1)}{P_{s' \sim \D_p}(s'|s)} - \frac{P(s'|s,a_2)}{P_{s' \sim \D_p}(s'|s)}} < \epsilon,
$$
giving us 
\begin{equation}
\label{eq:tv_distance}
\abs{P(s'|s,a_1) - P(s'|s,a_2)} < \epsilon.
\end{equation}

Next, let $\M$ be the MDP defined by $(\s, \A, P, R, \gamma)$, and let $\hat{\M}$ be the MDP defined by $(\s, \A, \hat{P}, R, \gamma)$, where
$$
\hat{P}(s'|s,a) =
\begin{cases}
\sum_{a' \in K} P(s'|s,a')\mu(a'|s) &,a \in K \\
P(s'|s,a) &,o.w.
\end{cases}
$$

For all $s \in \s, a \in \A$ we have that
\begin{align*}
&\abs{P(s'|s,a) - \hat{P}(s'|s,a)} \\
&\leq \sup_{a_1 \in K} \abs{P(s'|s,a_1) - \sum_{a' \in K} P(s'|s,a')\mu(a'|s)} \\
& = \sup_{a_1 \in K} \abs{ \sum_{a' \in K} \pth{P(s'|s,a_1) - P(s'|s,a')}\mu(a'|s)} \\
& \leq \sup_{a_1 \in K} \sum_{a' \in K} \abs{P(s'|s,a_1) - P(s'|s,a')}\mu(a'|s)
< \epsilon,
\end{align*}
where in the last step we used Equation \ref{eq:tv_distance} and the fact that $\sum_{a' \in K}\mu(a'|s) =1$. For the remainder of the proof we will use the following result proven in \cite{thm1_ref}:
\begin{lemma}
\label{lemma:abbeel}
Let $\M, \hat{\M}$ be MDPs as defined above. If
$$
\sum_{s'} \abs{P(s'|s,a) - \hat{P}(s'|s,a)} < \epsilon, \forall s,a
$$
then
$$
\sum_{s_t} \abs{P(s_t|s_0) - \hat{P}(s_t|s_0)} < t\epsilon.
$$
\end{lemma}

By definition of $\hat{\M}$ and $\pi_K$ we have that $\hat{P}^\pi = P^{\pi_K}$. Then, for all $s \in \s$
\begin{align*}
&\abs{V^\pi(s) - V^{\pi_K}(s)} \\
&= \abs{\E_{P^\pi} \pth{\sum_{t=0}^\infty \gamma^t r(s_t) \middle| s_0 = s} -
\E_{\hat{P}^\pi} \pth{\sum_{t=0}^\infty \gamma^t r(s_t) \middle| s_0 = s}} \\
& = \abs{\sum_{t=0}^\infty \gamma^t \bpth{\E_{P^\pi} \pth{r(s_t) | s_0 = s} - \E_{\hat{P}^\pi} \pth{r(s_t) | s_0 = s}}} \\
& \leq \sum_{t=0}^\infty \gamma^t \abs{\E_{P^\pi} \pth{r(s_t) | s_0 = s} - \E_{\hat{P}^\pi} \pth{r(s_t) | s_0 = s}}.
\end{align*}
Writing the above explicitly we get
\begin{equation}
\abs{V^\pi(s) - V^{\pi_K}(s)} \leq
\sum_{t=0}^\infty \gamma^t \abs{\sum_{s} r(s) \pth{P(s_t|s_0 = s) - \hat{P}(s_t|s_0 = s)}}.
\label{eq:ineq1}
\end{equation}
Next denote the sets
\begin{align*}
&Q_t = \braces{(s_t) : P(s_t| s_0 = s) > \hat{P}(s_t | s_0 = s) }, \text{and} \\
&Q_t^c = \braces{(s_t) : P(s_t| s_0 = s) \leq \hat{P}(s_t| s_0 = s) }.
\end{align*}
Then,
\begin{align*}
&\sum_{s_t} r(s_t) \pth{P(s_t|s_0 = s) - \hat{P}_t(s_t,a)} \\
&= \sum_{s_t \in Q_t} r(s_t) \pth{P(s_t|s_0 = s) - \hat{P}(s_t|s_0 = s)} + \sum_{s_t \in Q_t^c} r(s_t) \pth{P(s_t|s_0 = s) - \hat{P}(s_t|s_0 = s)} \\
& \leq \sum_{s_t \in Q_t} r(s_t)\pth{P(s_t|s_0 = s) - \hat{P}(s_t|s_0 = s)} \\
&= \sum_{s_t \in Q_t} r(s_t)\abs{P(s_t|s_0 = s) - \hat{P}(s_t|s_0 = s)} \\
& \leq \sum_{s_t} r(s_t) \abs{P(s_t|s_0 = s) - \hat{P}(s_t|s_0 = s)} \\
& \leq \sum_{s_t} \abs{P(s_t|s_0 = s) - \hat{P}(s_t|s_0 = s)} .
\end{align*}
By switching the roles $P_t$ and $\hat{P}_t$ we get the opposite inequality:
\begin{equation*}
\sum_{s_t} r(s_t) \pth{\hat{P}(s_t|s_0 = s) - P(s_t|s_0 = s)} 
\leq \sum_{s_t} \abs{\hat{P}(s_t|s_0 = s) - P(s_t|s_0 = s)}.
\end{equation*}
Overall,
\begin{equation}
\sum_{s_t} r(s_t) \abs{\hat{P}(s_t|s_0 = s) - P(s_t|s_0 = s)} 
\leq \sum_{s_t} \abs{\hat{P}(s_t|s_0 = s) - P(s_t|s_0 = s)}.
\label{eq:ineq2}
\end{equation}
Plugging (\ref{eq:ineq2}) in (\ref{eq:ineq1}) and using Lemma \ref{lemma:abbeel} we get
\begin{align*}
&\abs{V^\pi(s) - V^{\pi_K}(s)} 
\leq \sum_{t=0}^\infty \gamma^t  \sum_{s_t} \abs{\hat{P}(s_t|s_0 = s) - P(s_t|s_0 = s)} 
 \leq \sum_{t=0}^\infty \gamma^t t \epsilon.
\end{align*}

The proof follows immediately due to
$$
\sum_{t=0}^\infty \gamma^t t = \gamma\frac{1 + 4\gamma + \gamma^2}{\pth{1-\gamma}^4} \leq \frac{6\gamma}{\pth{1-\gamma}^4}
$$
for $\abs{\gamma} < 1$.

\subsection{Experimental Details}

\subsubsection{Drawing}

We used the \textit{Quick, Draw!} \cite{quickdraw} dataset, consisting of 50 million drawings ranging over 345 categories. While the raw data describes strokes as relative positions of a user's pen
on the canvas, we simplified the action space to include four primitive actions: left, right, up, down, i.e., movement of exactly one pixel in each of the directions. This was done by connecting lines between any two consecutive points in a stroke, and ignoring any drawings consisting of multiple strokes. Finally, we defined our action space to be any sequence of length $k$ primitive actions $(a_1, . . . , a_k)$, which we will hereinafter denote as an action stroke.

For our task, we were concerned with drawing a square. We used the 70,000 human-drawn squares in the ``square" category of the \textit{Quick, Draw!} dataset. Scanning the corpus of square drawings we trained Act2Vec on 12 types of actions strokes: 
\begin{enumerate}
\item Strokes of length 20 pixels in each of the axis directions: Left$\times 20$, Right$\times 20$, Up$\times 20$, Down$\times 20$.
\item Strokes of length 20 pixels consisting of corners in each of the possible directions: Left$\times 10$+Up$\times 10$, Up$\times 10$+Right$\times 10$, Right$\times 10$+Down$\times 10$, Down$\times 10$+Left$\times 10$, Right$\times 10$+Up$\times 10$, Up$\times 10$+Left$\times 10$, Left$\times 10$+Down$\times 10$, Down$\times 10$+Right$\times 10$. 
\end{enumerate}

Act2Vec was trained with SGNS over $50$ epochs using a window size of $w=2$, embedding dimension $d=5$, and $k=5$ negative samples.

\begin{algorithm}[tb!]
   \caption{$Q$-Embedding with $k$-Exp}
   \label{alg:q-embedding}
\begin{algorithmic}
	\STATE Create Act2Vec Embeddings $\mathcal{E}$ for actions in $\A$
	\STATE Create $k$ clusters $\{C_i\}_{i=1}^k$ based on $\mathcal{E}$ using $k$-means
   \STATE Initialize replay memory $\mathcal{R}$
   \FOR{episode $=1$ {\bfseries to} $m$}
   \STATE Reset simulator
   \FOR{$t=1$ {\bfseries to} $T$}
   \STATE $Q(s_t, a_t^i ; \theta) \gets \psi_\theta(\mathcal{E}(a_t^i))^T \phi_\theta(s_t)$
   \STATE Sample $U$ uniformly in $[0,1]$
   \IF{$U < \epsilon$}
   \STATE Sample $i$ uniformly in $\{1, \hdots, k\}$
   \STATE Sample action $a$ uniformly in $C_i$
   \ELSE
   \STATE Choose action $a$ which maximizes $Q(s_t, a; \theta)$
   \ENDIF
   \STATE Execute action $a$ in simulator.
   \STATE Observe reward $r_t$ and next state $s_{t+1}$
   \STATE Store transition $(s_t, a_t, r_t, s_{t+1})$ in $\mathcal{R}$.
   \STATE Sample random minibatch of transitions $(s_j, a_j, r_j, s_{j+1})$ from $\mathcal{R}$
   \STATE $y_j \gets 
   \begin{cases} 
   r_j & \text{$s_{j+1}$ is terminal} \\
   r_j + \gamma \max_{a' \in \A} Q(s_{j+1}, a' ; \theta) & \text{o.w.}
   \end{cases}$
   \STATE Perform gradient descent step on $(y_j - Q(s_j, a_j ; \theta))^2$ w.r.t. network parameters $\theta$
   \ENDFOR
   \ENDFOR
\end{algorithmic}
\end{algorithm}

\samepage{The reward function was given by ${R = \frac{\sum_{i=1}^4 \min \pth{W, l_i}}{4W}}$, where $l_i$ denotes the length of the $i^{th}$ side of the drawn shape, and $W$ is the desired length of each side. When the drawn shape had more or less than four sides, a reward of $-0.1$ was given. In our case we tested values of $W=40, 60$.}

The state was represented as the sum of embeddings of previously taken actions. That is,
$$
s_t = \sum_{i=0}^{t-1} \mathcal{E}(a_i),
$$
where $\mathcal{E}$ are the used embeddings. We tested four types of embeddings: Act2Vec, Normalized Act2Vec (by $l_2$ norm), one-hot embedding, and random embedding. The one-hot embedding is defined as the unit vector $e_i$ of length $\abs{\A}$, with zeros everywhere except for position $i$, corresponding to action $a_i$. Random embeddings were initialized uniformly in $[0,1]^d$.

The agent was trained using Proximal Policy Optimization (PPO) \cite{ppo} over $1$ million iterations for the case of $W=40$ and $10$ million for $W=60$ using $256$ asynchronous threads. PPO was run using the following parameters: learning rate = $0.0005$, $\lambda = 0.95$, $\gamma = 0.99$, $3$ experience buffer passes, entropy coefficient = $0.01$, clipping range = $0.1$. The value and policy networks were both represented as feedforward neural networks with two hidden layer with $64$ units each. Each of the state representation methods were tested for a set of 15 trials, except for random embeddings, which were tested for a set of 100 trials. A different random embeddings was sampled for each tested trial.

\begin{figure}[t!]
\vskip 0.2in
\begin{center}
\begin{subfigure}{0.4\textwidth}
\includegraphics[width=\linewidth]{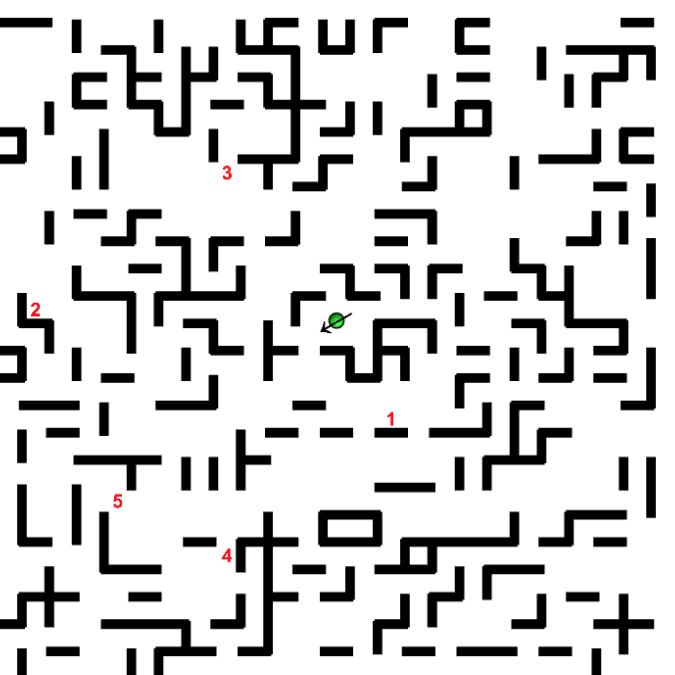}
\caption{}\label{fig:2d_env}
\end{subfigure}
\hskip 0.5in
\begin{subfigure}{0.4\textwidth}
\includegraphics[width=\linewidth]{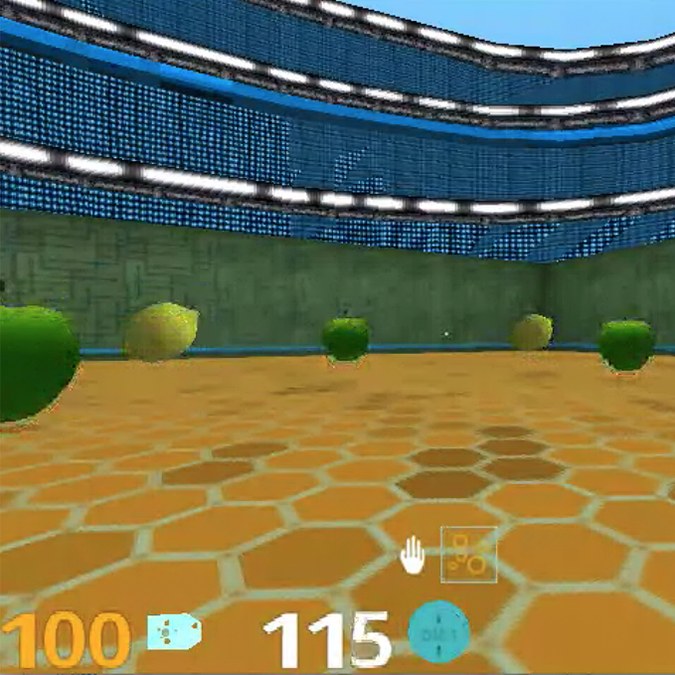}
\caption{}\label{fig:3d_env}
\end{subfigure}
\caption{Images show the 2d navigation (a) and 3d Seek-Avoid (b) environments used in Section \ref{sec:navigation}. Action embeddings learned in the 2d environment were used as knowledge transfer to improve the solution of the task in the 3d environment. }
\label{fig:navig_envs}
\end{center}
\vskip -0.2in
\end{figure}

\subsubsection{Navigation}

We start by describing the 2d environment in which the action embeddings where learned. An image of the environment can be seen in Figure \ref{fig:2d_env}. The environment was divided to a 25x25 grid, in which 300 walls were randomly placed as well as 5 numbers relating to goals. The player was initialized in the center of the grid. The player's movement was not constrained to the grid, but rather a more continuous movement -  similar to that of the Seek-Avoid environment. The player was given the following three actions: Move forward one unit, rotate left 25 degrees, and rotate right 25 degrees. The player was given the task to move from its initial position and reach each of the goal numbers (in order). 

Actions of a demonstrator player were recorded for different initializations of the 2d environment. Act2Vec was trained over a corpus of 3000 actions. Act2Vec was trained with SGNS over $50$ epochs using different window sizes of $w=2,3$, embedding dimensions $d=3,5$, and $k=5$ negative samples.

The 3d environment used to test the DQN agent was the Seek-Avoid lab \cite{deepmind_lab}. An image of the environment can be see in in Figure \ref{fig:3d_env}. The input image was resized to a resolution of $80\times80$ pixels with $3$ channels (RGB). All experiments used the same state representation: a convolutional neural network with 3 layers: (1) 2d convolution with 8 channels, window size of $3\times3$ and stride of $2$, (2) 2d convolution with 16 channels, window size of $3 \times3$, \mbox{and stride of 2}, and (3) fully connected layer with $128$ hidden features, $\phi_\theta(s)$. For the case of the traditional DQN, the network output was $w_a^T \phi_\theta(s)$, where $w_a$ were the network's parameters learned for each action. For the case of $Q$-Embedding, we used another network to approximate $w_a$ as $\psi_\theta(\mathcal{E}(a))$, where $\mathcal{E}(a)$ is the embedding of action $a$, and $\psi_\theta$ is a feedforward neural network with one hidden layer of 64 units and $tanh$ activation function, and 128 outputs. In this case, the network output was given by $\psi_\theta(\mathcal{E}(a))^T \phi_\theta(s)$, for each of the actions.

The action space consisted of: move forward one unit, turn view 25 degrees to the right, and turn view 25 degrees to the left. Every action was repeated for a sequence of 10 frames.  In the case of action sequences, every action in the sequence was repeated for 10 frames, i.e., a sequence of $k$ actions was executed for $10k$ frames.

The agent used a replay memory of one million transitions and a training batch of $64$ transitions. The agent was trained over 60000 iterations with a learning rate of 0.00025, and discount factor of $\gamma = 0.99$. For uniform and $k$-Exp exploration decaying $\epsilon$ was used, starting at a value of $1$ and decaying linearly to a value of $0.1$ once reached to iteration $19800$.

\subsubsection{Q-Embedding Algorithm}

The $Q$-Embedding with $k$-Exp algorithm is presented in Algorithm \ref{alg:q-embedding}. An NLP  version of this algorithm (with uniform exploration) has already been proposed in \cite{nlp_rl}.

\end{document}